\documentclass[runningheads]{llncs}
\usepackage{graphicx}
\usepackage{comment}
\usepackage{amsmath,amssymb} %
\usepackage{color}
\usepackage{sidecap}
\sidecaptionvpos{figure}{c}
\usepackage{array,multirow}
\usepackage{multicol}
\usepackage{tablefootnote}
\usepackage{hyperref}

\usepackage[dvipsnames]{xcolor,colortbl}
\usepackage{todonotes}

\begin{document}

\newcommand{\cam}[1]{\textcolor{Green}{\textbf{Camilo: #1}}}
\newcommand{\anelise}[1]{\textcolor{Orange}{\textbf{Anelise: #1}}}

\newcommand{\parboxvpadding}{\vspace{.1cm}}
\newcommand{\norm}[1]{\left\lVert#1\right\rVert}

\pagestyle{headings}
\mainmatter
\def\ECCVSubNumber{2535}  %

\title{Multimodal Memorability: Modeling Effects of Semantics and Decay on Video Memorability} %

\newcommand\edit[1]{\textcolor{black}{#1}}

\titlerunning{Multimodal Memorability}
\author{Anelise Newman* \and
Camilo Fosco* \and 
Vincent Casser \and
Allen Lee \and
Barry McNamara \and
Aude Oliva}
\authorrunning{A. Newman et al.}
\institute{Massachusetts Institute of Technology \\
}
\maketitle

\renewcommand{\thefootnote}{\fnsymbol{footnote}}
\footnotetext[1]{Equal contribution.}
\renewcommand{\thefootnote}{\arabic{footnote}}

\begin{abstract}
   A key capability of an intelligent system is deciding when events from past experience must be remembered and when they can be forgotten. 
   Towards this goal, we develop a predictive model of human visual event memory and how those memories decay over time. 
   We introduce \textit{Memento10k}, a new, dynamic video memorability dataset containing human annotations at different viewing delays.
   Based on our findings we propose a new mathematical formulation of memorability decay, resulting in a model that is able to produce the first quantitative estimation of how a video decays in memory over time. 
   In contrast with previous work, our model can predict the probability that a video will be remembered at an arbitrary delay.
   Importantly, our approach \edit{combines} visual and semantic information \edit{(in the form of textual captions)} to fully represent the meaning of events. 
   Our experiments on two video memorability benchmarks, including Memento10k, show that our model significantly improves upon the best prior approach (by 12\% on average). 
\end{abstract}

\section{Introduction}

Deciding which moments from past experience to remember and which ones to discard is a key capability of an intelligent system. 
The human brain is optimized to remember what it deems to be important and forget what is uninteresting or redundant.
Thus, human memorability is a useful measure of what content is interesting and likely to be retained by a human viewer.
If a system can predict which information will be highly memorable, it can evaluate the utility of incoming data and compress or discard what is deemed to be irrelevant. 
It can also filter to select the content that will be most memorable to humans, which has potential applications in design and education.

However, memorability of dynamic events is challenging to predict because it depends on many factors. First, different visual representations are forgotten at different rates: while some events persist in memory even over long periods, others are forgotten within minutes \cite{Bainbridge2013,Lore2017memorability,isola2014pami,Konkle2010SceneMemory}. 
This means that the probability that someone will remember a certain event varies dramatically as a function of time, introducing challenges in terms of how memorability is represented and measured.
Second, memorability depends on both visual and semantic factors. In human cognition, language and vision often act in concert for remembering an event.
Events described with richer and more distinctive concepts are remembered for longer than events attached to shallower descriptions, and certain semantic categories of objects or places are more memorable than others \cite{isola2014pami,Konkle2010Concepts}.

\begin{figure}[t]
\centering
  \includegraphics[width=1\columnwidth]{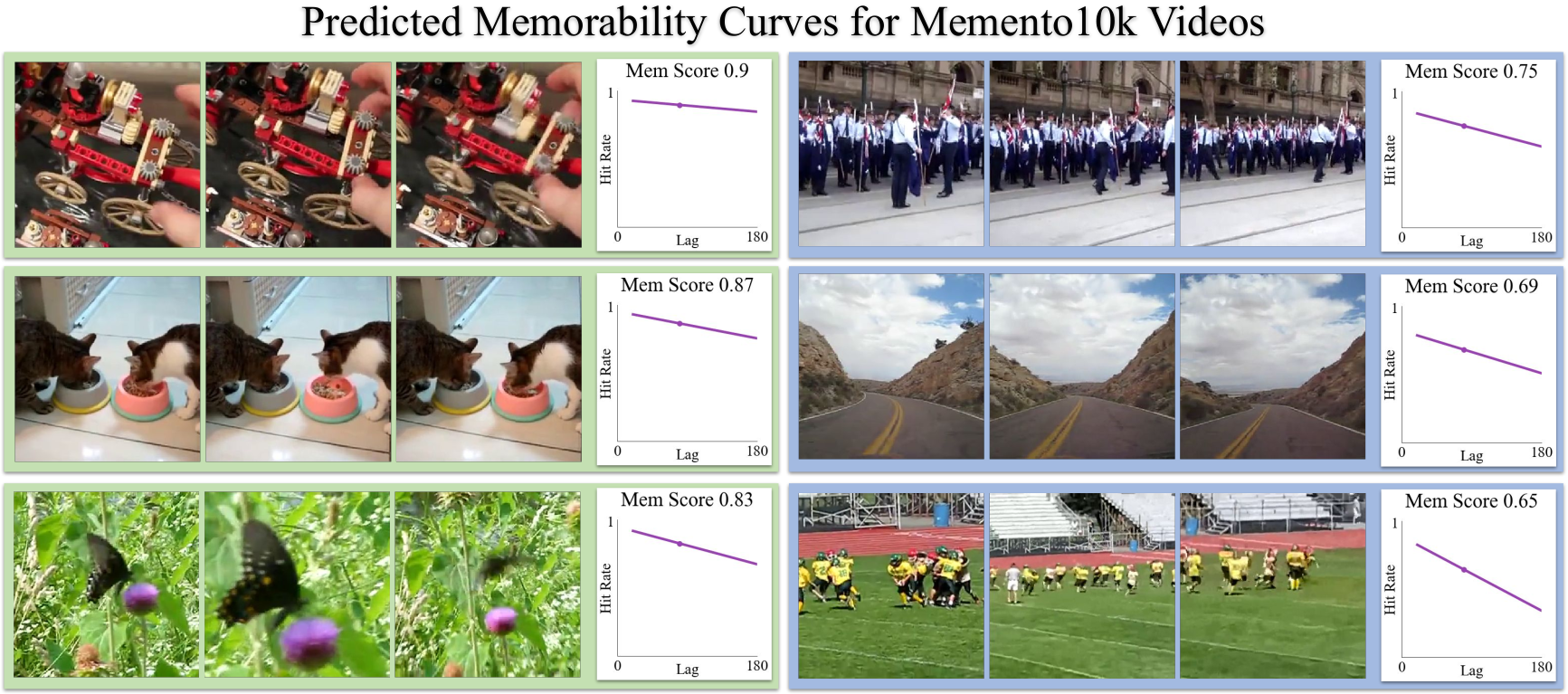}
  \caption{\textbf{How do visual and semantic features impact memory decay over time?} 
  We introduce a multimodal model, SemanticMemNet, that leverages visual and textual information to predict the memorability decay curve of a video clip. 
  We show predictions on videos from Memento10k, our new video memorability dataset. 
  SemanticMemNet is the first model to predict a decay curve that represents how quickly a video falls off in memory over time.
  } ~\label{fig:hero_fig}
\end{figure}

In this paper, we introduce a new dataset and a model for video memorability prediction that address these challenges\footnote{Dataset, code, and models can be found at our website: \url{http://memento.csail.mit.edu/}}. 
Memento10k, the most dynamic video memorability dataset to date, contains both human annotations at different viewing delays and human-written captions, making it ideal for studying the effects of delay and semantics on memorability.
Based on this data, we propose a mathematical formulation of memorability decay that allows us to estimate the probability that an event will be remembered at any point over the first ten minutes after viewing. 
We introduce SemanticMemNet, a multimodal model \edit{that relies on visual and semantic features to predict} the decay in memory of a short video clip (Fig. \ref{fig:hero_fig}). 
SemanticMemNet is the first model that predicts the entire memorability decay curve, which allows us to estimate the probability that a person will recall a given video after a certain delay.
We also enhance our model's features to include information about video semantics by jointly predicting verbal captions. 
SemanticMemNet achieves state-of-the-art performance on video memorability prediction on two different \edit{video memorability} baselines.

To summarize, our key contributions are:
\begin{itemize}

    \item We present a new \textbf{multi-temporal, multimodal memory dataset} of 10,000 video clips, \emph{Memento10k}. With \edit{over} 900,000 human memory annotations at different delay intervals and 50,000 captions describing the events, it is the largest repository of dynamic visual memory data.
    \item We propose a \textbf{new mathematical formulation for memorability decay} which estimates the decay curve of human memory for individual videos.
    \item We introduce \textbf{SemanticMemNet}, a model capitalizing on learning visual and semantic features to predict both the memorability strength and the memory decay curve of a video clip.

\end{itemize}

\section{Related Work}

\noindent \textbf{Memorability in Cognitive Science.} Four landmark results in cognitive science inspired our current approach. 
First, memorability is an \textit{intrinsic} property of an image: people are remarkably consistent in which images they remember and forget \cite{Bainbridge2013,bylinskii2015intrinsic,Lore2017memorability,isola2014pami,isola2011cvpr,Rust2019,khosla2015,Mohsenzadeh2019}. 
Second, semantic features, such as a detailed conceptual representation or a verbal description, boost visual memory and aid in predicting memorability \cite{Konkle2010Concepts,Koutstaal2003,Vogt2007,Wiseman1974}. 
Third, most stimuli decay in memory, with memory performances falling off predictably over time \cite{Lore2017memorability,isola2014pami,Konkle2010SceneMemory,vo2017memorability}. 
Finally, the classical old-new recognition paradigm (i.e. people press a key when they recognize a repeated stimulus in a sequence) allows researchers to collect objective measurements of human memory at a large scale \cite{brady2008visual,isola2011cvpr} and variable time scales, which we draw on to collect our dataset.

\noindent \textbf{Memorability in Computer Vision.} 
\edit{The} intrinsic nature of visual memorability means that visual stimuli themselves contain visual and semantic features that can be captured by machine vision. 
For instance, earlier works \cite{isola2011nips,isola2014pami,isola2011cvpr} pointed to content of images that were predictive of their memorability (i.e. people, animals and manipulable objects are memorable, but landscapes are often forgettable).  
Later work replicated the initial findings and extended \edit{the memorability prediction task} to many photo categories \cite{Akagunduz2019,baveye2016,dubey2015,fajtl2018,khosla2012,Perera2019,zarezadeh2017}, faces \cite{Bainbridge2013,khosla2013,Sidorov2019}, visualizations \cite{Borkin2016,Borkin2013} and videos \cite{Cohendet2019ICCV,cohendet2018annotating,shekhar2017}. 

The development of large-scale image datasets augmented with memorability scores \cite{khosla2015} allowed convolutional neural networks to predict image memorability at near human-level consistency \cite{Akagunduz2019,baveye2016,fajtl2018,khosla2015,zarezadeh2017} and even generate realistic memorable and forgettable photos \cite{ganalyzeiccv2019,Sidorov2019}. 
However, similar large-scale work on video memorability prediction has been limited. 
In the past year, Cohendet et al. introduced a video-based memorability dataset \cite{cohendet2018} that is the only other large benchmark comparable to Memento10k, and made progress towards building a predictive model of video memorability \cite{Cohendet2019ICCV,cohendet2018annotating}.
Other works on video memorability have largely relied on smaller datasets collected using paradigms that are more challenging to scale. 
\cite{Han2015} collected memorability and fMRI data on 2400 video clips and \edit{aligned} audio-visual features with brain data \edit{to improve} prediction. 
\cite{shekhar2017} used a language-based recall task to collect memorability scores as a function of response time for 100 videos. They find that a combination of semantic, spatio-temporal, saliency, and color features can be used to predict memorability scores. 
\cite{cohendet2018annotating} collected a long-term memorability dataset using clips from popular movies that people have seen before. 

Past work has confirmed the usefulness of semantic features for predicting memorability \cite{Cohendet2019ICCV,hammad2018,shekhar2017}.
Work at the intersection of Computer Vision and NLP has aimed to bridge the gap between images and text by generating natural language descriptions of images using encoder-decoder architectures (e.g. \cite{karpathy_deep_alignments_2014,s2vt_2015,show_attend_tell_2015}) or by creating aligned embeddings for visual and textual content \cite{beans_in_burgers_2018,devise_2013,kiros_visual_semantic_embedding_2014}. Here, we experiment with both approaches in order to create a memorability model that jointly learns to understand visual features and textual labels.

\section{Memento10k: A Multimodal Memorability Dataset} \label{sec:memento_video_mem_game}

\begin{figure}[h]
\centering
  \includegraphics[width = 1\columnwidth]{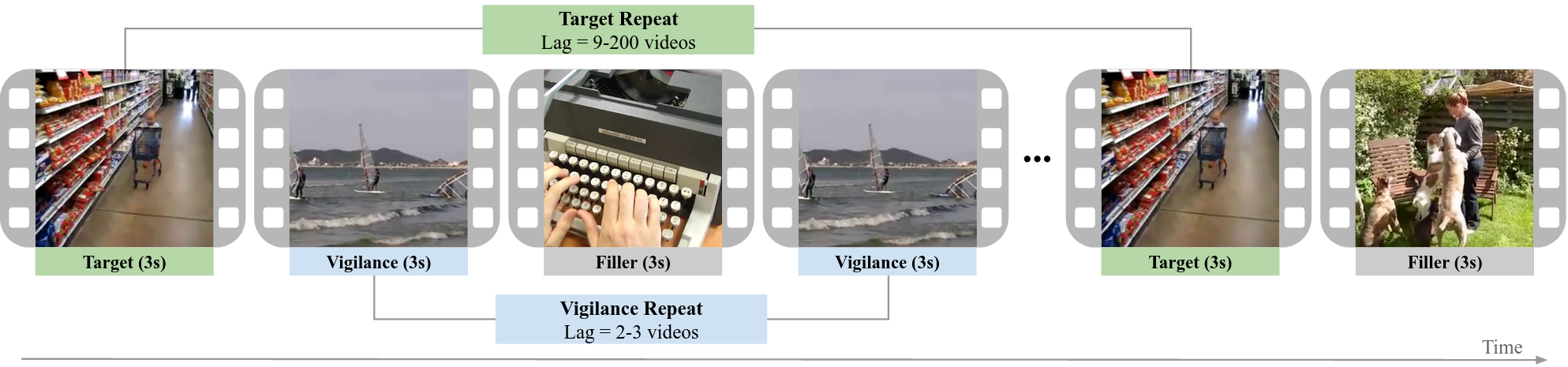}
  \caption{Task flow diagram of \textit{The Memento Video Memory Game}. Participants see a continuous stream of videos and press the space bar when they see a repeat. %
  } ~\label{fig:memento_task_diagram}
\end{figure}

Memento10k \edit{focuses on both} the visual and semantic underpinnings of video memorability. 
The dataset contains 10,000 video \edit{clips} augmented with memory scores, action labels, and textual descriptions (five human-generated captions per video). 
Importantly, our memorability annotations occur at presentation delays ranging from several seconds to ten minutes, which, for the first time, allows us to model how memorability falls off with time.

\begin{figure}[t]
\centering
  \includegraphics[width = 0.95\columnwidth]{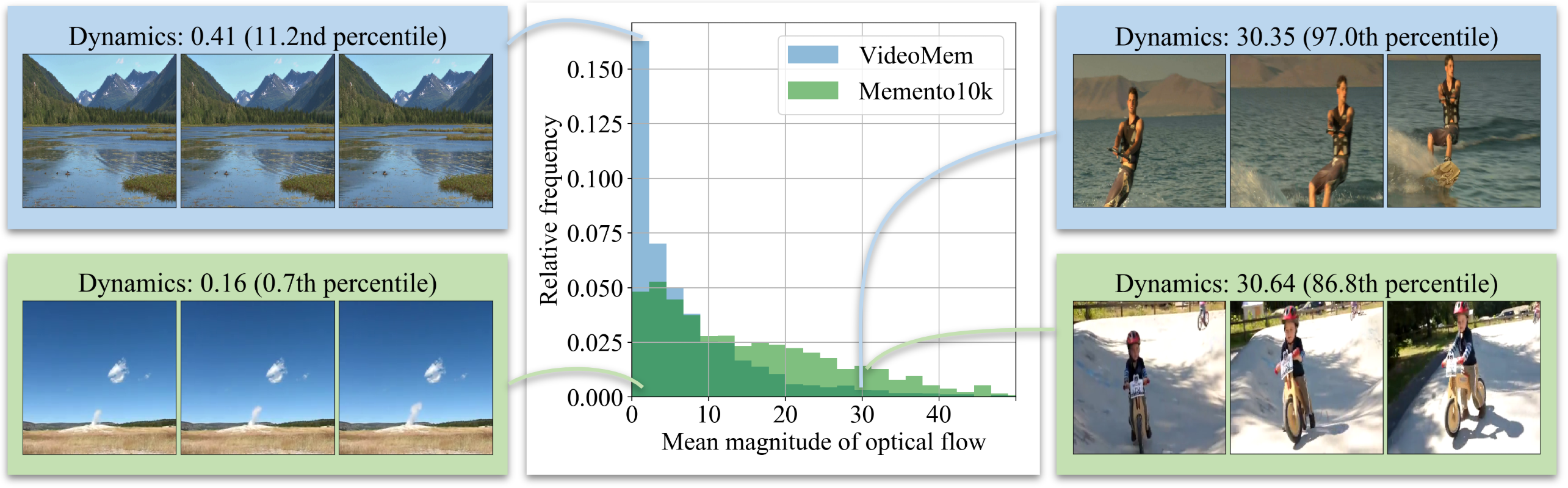}
  \caption{\textbf{Video distribution by level of motion for VideoMem and Memento10k.} The motion metric for each video is calculated by averaging optical flow magnitude over the entire video. A high percentage of videos in the VideoMem dataset are almost static, while Memento10k is much more balanced. We show select examples with low and high levels of motion.}
 ~\label{fig:dyn_fig}
\end{figure}

\begin{figure}[t]
\centering
  \includegraphics[width = 1\columnwidth]{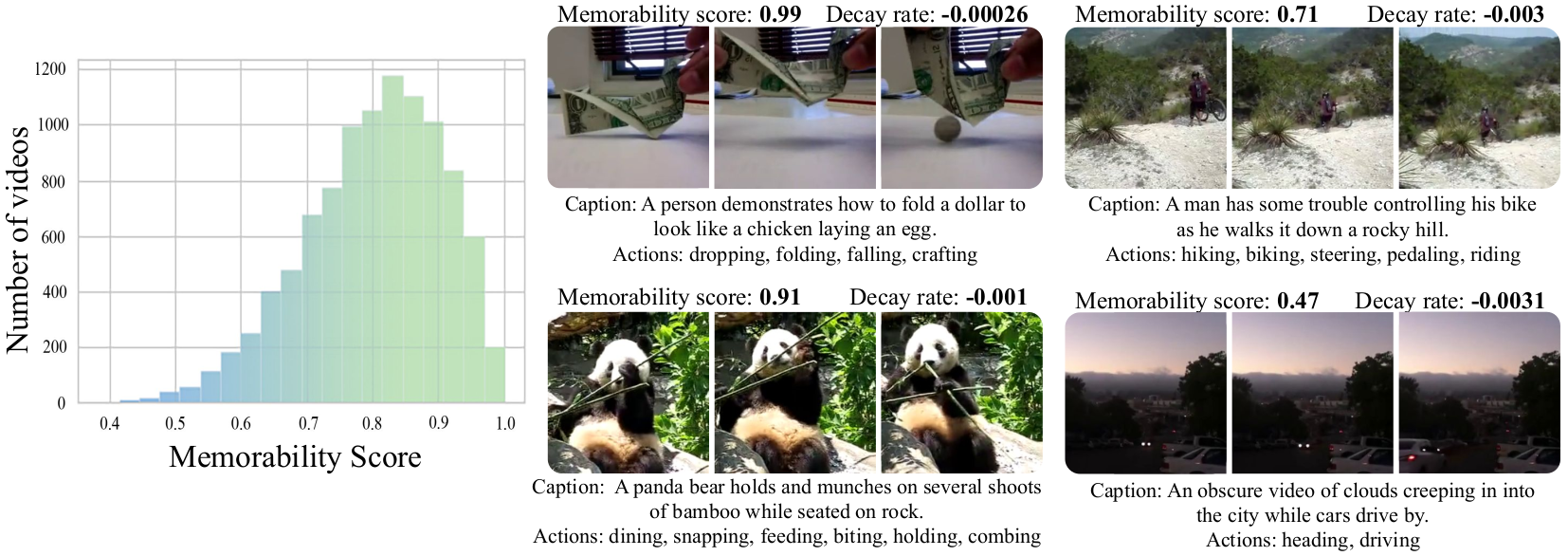}
  \caption{
  \textbf{The Memento10k dataset} contains the memorability scores, alpha scores (decay rates), action labels, and five unique captions for 10,000 videos. 
  \textbf{Left:} The distribution of memorability scores over the entire dataset. \textbf{Right:} Example clips from the Memento10k dataset along with their memorability score, decay rate, actions, and an example caption.}
 ~\label{fig:memento_dataset_figure}
\end{figure}

\noindent \textbf{The Memento Video Memory Game.} 
In our \edit{experiment}, crowdworkers from Amazon's Mechanical Turk (AMT) watched a continuous stream of three-second video clips and were asked to press the space bar when they saw a repeated video.
Importantly, we varied the lag for repeated videos (the number of videos between the first and the second showing of a repeat), which allowed us to study the evolution of memorability over time. 
Our policies came from \cite{isola2011cvpr,khosla2015} and a summary of paradigm details can be found in the supplement. 

\edit{Three seconds, the length of a Memento clip, is about the average duration of human working memory \cite{Baddeley_working_memory,barrouillet2004timeconstraints} and most human visual memory performances plateau at three seconds of exposure \cite{brady2008visual,brady2013fidelity}. 
This makes three seconds a good ``atomic'' length for a single item held in memory.
Previous work has shown that machine learning models can learn robust features even from short clips \cite{monfort2019moments}.}

\noindent \textbf{Dynamic, In-The-Wild Videos.} \label{section:memento10k-dynamic}
Memento10k is composed of natural videos scraped from the Internet
\footnote{The Memento videos have partial overlap with the Moments in Time \cite{monfort2019moments} dataset.}.
To limit our clips to non-artificial scenes with everyday context, we asked crowdworkers whether each clip was a ``home video" and discarded videos that did not meet this criterion.
After removing clips that contained undesirable properties (i.e. watermarks), we were left with 10,000 ``clean'' videos, which we break into train (\edit{7000}), validation (\edit{1500}), and test (\edit{1500}) sets. 

The Memento10k dataset 
is a significant step towards understanding memorability of real-world events. 
First, it is the most dynamic memorability dataset to date with videos containing a variety of motion patterns including camera motion and moving objects. 
The mean magnitude of optical flow in Memento10k is nearly double that of VideoMem \cite{Cohendet2019ICCV} (\edit{approximately} 15.476 for Memento vs. 7.296 for VideoMem, see Fig. \ref{fig:dyn_fig}), whose clips tend to be fairly static. 
Second, Memento10k's diverse, natural content enables the study of memorability in an everyday context: Memento10k was compiled from in-the-wild amateur videos like those found on social media or real-life scenes, while VideoMem is composed of professional footage. 
Third, Memento10k's greater number of annotations (90 versus 38 per video in VideoMem), spread over lags of 30 seconds to 10 minutes, leads to higher ground-truth human consistency and allows for a robust estimation of a video's decay rate.  
Finally, we provide more semantic information such as action labels as well as 5 detailed captions per video. 
In this paper, we use both benchmarks to evaluate the generalization of our model.

\begin{figure}[t]
    \centering   
    
    \includegraphics[width=1\columnwidth]{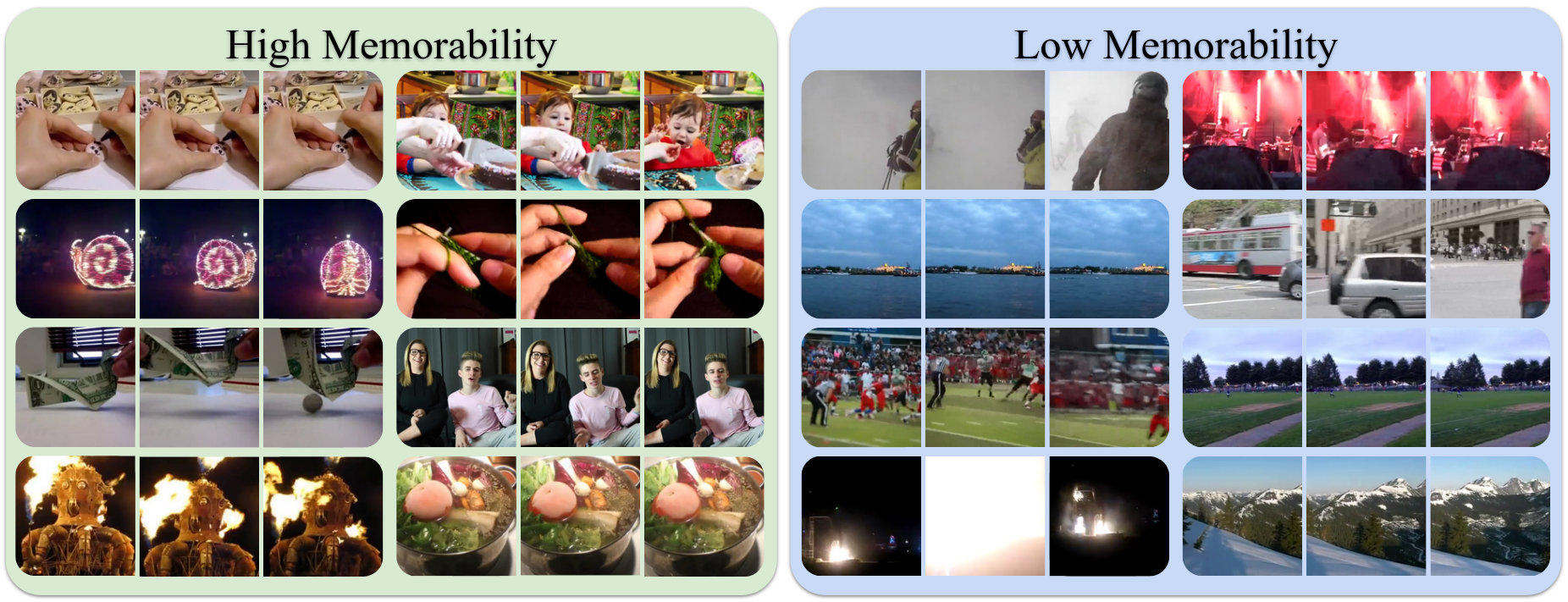}
  
    \caption{
    \textbf{Examples of high- and low-memorability video clips.} 
    Videos involving people, faces, hands, man-made spaces, and moving objects are in general more memorable, while clips containing distant/outdoor landscapes or dark, cluttered, or static content tend to be less memorable.
    }
    \label{fig:mem_examples}
\end{figure}

\noindent \textbf{Semantic Annotations.} We augment our dataset with captions, providing a source of rich textual data that we can use to relate memorability to semantic concepts (examples in Fig. \ref{fig:memento_dataset_figure}).
We asked crowdworkers to describe the events in the video clip in full sentences and we manually vetted the captions for quality and corrected spelling mistakes. 
\edit{Each video has 5 unique captions from different crowdworkers.
More details on caption collection are in the supplement.}

\noindent \textbf{Human Results.} The Memento10k dataset contains over 900,000 individual annotations, making it the biggest memorability dataset to date. 
We measured human consistency \edit{of these annotations} following \cite{Cohendet2019ICCV,isola2011cvpr}: we randomly split our participant pool into two groups and calculate the Spearman's rank correlation between the memorability rankings produced by each group, where the rankings are generated by sorting videos by raw hit rate. 
The average rank correlation ($\rho$) over 25 random splits is \textbf{$0.73$} (compared to $0.68$ for images in \cite{khosla2015}, and $0.616$ for videos in \cite{Cohendet2019ICCV}). 
This high consistency between human observers confirms that videos have strong intrinsic visual, dynamic or semantic features that a model can learn from to predict memorability of new videos. 

Fig. \ref{fig:mem_examples} illustrates some qualitative results of our experiment. 
We see \edit{some similar patterns as with image memorability:}
memorable videos tend to contain saturated colors, people and faces, manipulable objects, and man-made spaces, while less memorable videos are dark, cluttered, or inanimate. 
Additionally, videos with interesting motion patterns can be highly memorable whereas static videos often have low memorability.

\section{Memory Decay: A Theoretical Formulation }

Most memories decay over time. In psychology this is known as the forgetting curve, which estimates how the memory of an item naturally degrades. 
Because Memento10k's memorability annotations occur at lags of anywhere from 9 videos (less than 30 seconds) to 200 videos (around 9 minutes), we have the opportunity to calculate the strength of a given video clip's memory at different lags.  

\begin{figure}[t]
    \centering
    \includegraphics[width=.75\columnwidth]{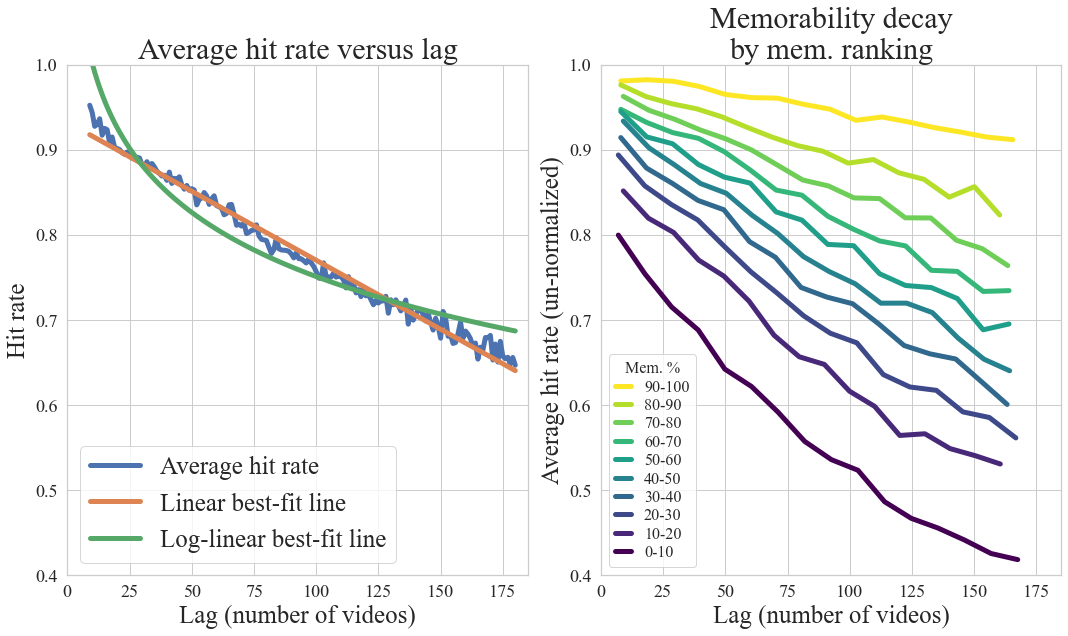}
    \caption{Our data suggests a \textbf{memory model where each video decays linearly} in memory according to an individual decay rate $\alpha^{(v)}$. 
    \textbf{Left:} A linear trend is a better approximation for our raw data \edit{($r=-0.991$)} than a log-linear trend \edit{($r=-0.953$)}. 
    \textbf{Right:} We confirm our assumption that $\alpha$ varies by video by grouping videos into deciles based on their normalized memorability score and plotting group average hit rate as a function of lag. Videos with lower memorability show a faster rate of decay. 
    }
\label{fig:mem_score_calculation}
\end{figure}

A naive method for calculating a memorability score is to simply take the video's target hit rate, or the fraction of times that the repeated video was correctly detected. 
However, since we expect a video's hit rate to go down with time, annotations at different lags are not directly comparable.  Instead, we derive an equation for how each video's hit rate declines as a function of lag.

First, for the lags tested in our study, we observe that hit rate decays linearly as a function of lag. This is %
notable because previous work on image memorability has found that images follow a log-linear decay curve \cite{Lore2017memorability,isola2014pami,vo2017memorability}. Fig. \ref{fig:mem_score_calculation} (left) shows that a linear trend best fits our raw annotations; \edit{this holds for videos across the memorability distribution (see the supplement for more details).}

Second, in contrast to prior work, we find that different videos decay in memory at different rates. 
Instead of assuming that all stimuli decay at one universal decay rate, $\alpha$, as in \cite{khosla2015}, we assume that each video decays at its own rate, $\alpha^{(v)}$. 
Following the procedure laid out in \cite{khosla2015}, we find a memorability score and decay rate for each video that approximates our annotations. 
We define the memorability of video $v$ as $m_T^{(v)} = \alpha^{(v)} T + c^{(v)}$, where $T$ is the lag (the interval in videos between the first and second presentation) and $c^{(v)}$ is the base memorability of the video. If we know $m_T^{(v)}$ and $\alpha^{(v)}$, we can then calculate the video's memorability at a different lag $t$ with $m_t^{(v)} = m_T^{(v)} + \alpha^{(v)} (t-T) ~\refstepcounter{equation}(\theequation)\label{eq1}$.

To obtain values for $m_T^{(v)}$ and $\alpha^{(v)}$, we minimize the L2 norm between the raw binary annotations from our experiment $x_j^{(v)}$, $j \in \{0, ..., n^{(v)}\}$ and the predicted memorability score at the corresponding lag, $m_t^{(v)}$. The error equation is:
\[
    E(\alpha^{(v)}, m_T^{(v)}) = 
    \sum_{j=1}^{n^{(v)}} \norm{x_j^{(v)} - m_t^{(v)} }_2^2 =  \sum_{j=1}^{n^{(v)}}\norm{x_j^{(v)} - \left[m_T^{(v)} + \alpha^{(v)} (t_j^{(v)}-T)\right] }_2^2
\]

We find update equations for $m_T^{(v)}$ and $\alpha^{(v)}$ by taking the derivative with respect to each and setting it to zero:
\small
\begin{align}
\alpha^{(v)} & \gets \frac{\frac{1}{n^{(v)}} \sum_{j=1}^{n^{(v)}} (t_j^{(v)}-T)\left[x_j^{(v)}-m_T^{(v)}\right]}{\frac{1}{n^(v)} \sum_{j=1}^{n^{(v)}} \left[t_j^{(v)}-T\right]^2} \quad \quad
 m_T^{(v)} & \gets \frac{1}{n^{(v)}} \sum_{j=1}^{n^{(v)}} \left[ x_j^{(v)} - \alpha^{(v)} (t_j^{(v)} - T) \right]
\end{align}
\normalsize

We initialize $\alpha^{(v)}$ to \edit{$-5\text{e}{-4}$} %
and $m_T^{(v)}$ to each video's mean hit rate. We set our base lag $T$ to $80$ and optimize for 10 iterations to produce $\alpha^{(v)}$ and $m_{80}^{(v)}$ for each video. We thus define a video's ``memorability score'', for the purposes of memorability ranking, as its hit rate at a lag of 80; however, we can use equation \ref{eq1} to calculate its hit rate at an arbitrary lag within the range that we studied.

Next, we validate our hypothesis that videos decay in memory at different rates. 
We bucket the Memento10k videos into 10 groups based on their normalized memorability scores and plot the raw data (average hit rate as a function of lag) for each group. 
Fig. \ref{fig:mem_score_calculation} (right) confirms that different videos decay at different rates.

\section{Modeling Experiments} 

In this section, we explore different architecture choices for modeling video memorability that take into consideration both visual and semantic features. 
We also move away from a conception of memorability as a single value, and instead predict the \emph{decay curve} of a video, resulting in the first model that predicts the raw probability that a video will be remembered at a particular lag. 

\noindent \textbf{Metrics.} 
A commonly-used metric for evaluating memorability is the Spearman rank correlation (RC) between the memorability ranking produced by the ground-truth memorability scores versus the predicted scores.
This is a popular metric \cite{Cohendet2019ICCV,isola2011cvpr,khosla2015} because memorability rankings are generally robust across experimental designs and choice of lag; however, it does not measure the accuracy of predicted memorability values.  %
As such, when evaluating the quality of the decay curve produced by our model, we will also consider $R^2$ for our predictions.

\subsection{Modeling Visual Features} \label{section:modeling-visual-features}

\def\arraystretch{1.05} %
\begin{table}[t]
    \centering
    \begin{tabular}{|c|c|c|c|}
        \cline{2-4}
          \multicolumn{1}{c|}{} 
          & \multicolumn{1}{c|}{\textbf{Approach}} 
          & \multicolumn{1}{c|}{\parbox[c]{3.0cm}{\parboxvpadding \centering  \textbf{RC - Memento10k (test set)} \parboxvpadding}}
          & \multicolumn{1}{c|}{\parbox[c]{2.8cm}{\parboxvpadding \centering  \textbf{RC - VideoMem (validation set)} \parboxvpadding}} \\
         \cline{2-4}
         \multicolumn{1}{c|}{} & Human consistency & 0.730  & 0.616 \\ 
         \hline 
         \multirow{4}{*}{(Sec. \ref{section:modeling-visual-features})}%
         & Flow stream only & 0.579 & 0.425\\ 
         & Frames stream only & 0.595 & 0.527\\
         & Video stream only & 0.596 & 0.492 \\
         & Flow + Frames + Video & 0.659 & 0.555\\
         \hline 
         \multirow{2}{*}{(Sec. \ref{section:modeling-semantic-features})}%
         & {\centering Video stream + captions} & 0.602 & 0.512 \\
         & {\centering Video stream + triplet loss} & 0.599 & - \\
         \hline
         \multicolumn{1}{c|}{} & \textbf{SemanticMemNet (ours)} & \textbf{0.663} & \textbf{0.556}  \\ 
         \cline{2-4}
    \end{tabular}
    \caption{\textbf{SemanticMemNet ablation study.} We experiment with different ways of incorporating visual and semantic features into memorability prediction. 
    We measure performance by calculating the Spearman's rank correlation (RC) of the predicted memorability rankings with ground truth rankings on both Memento10k and VideoMem. 
    }
    \label{tab:ablation_study}
\end{table}

\noindent \textbf{Baseline: Static Frames.} 
We evaluate the extent to which static visual features contribute to video memorability by training a network to predict a video's memorability from a single frame.
We first train an ImageNet-pretrained DenseNet-121 to predict image memorability by training on the LaMem dataset \cite{khosla2015}, then finetune on the video memorability datasets.
At test time, a video's memorability score is calculated by averaging predictions over every 4th frame (\edit{about} 22 frames total for Memento videos).

\begin{figure}
    \centering   
    \includegraphics[width=1\columnwidth]{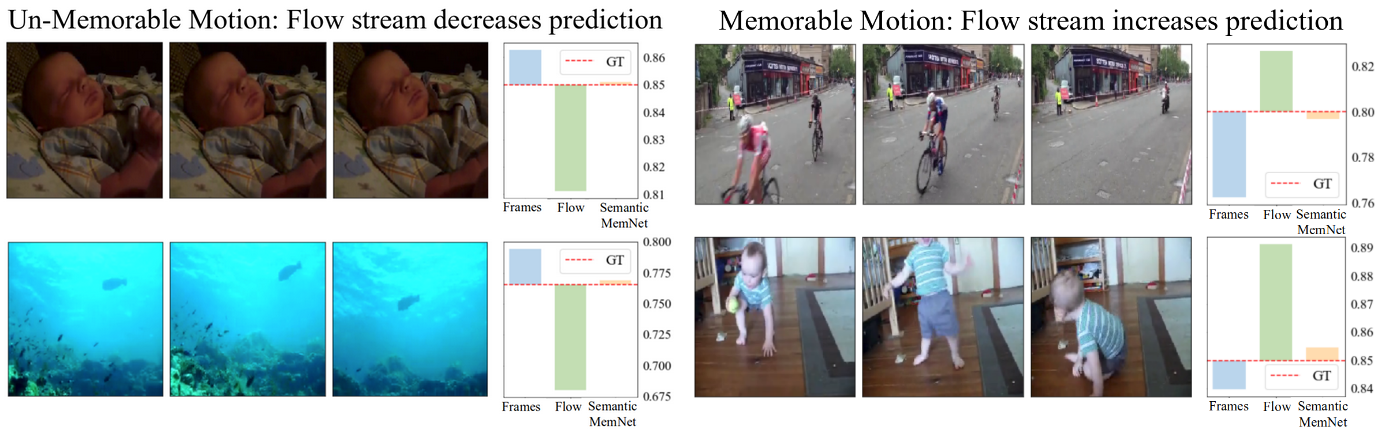}
    \caption{
    \textbf{Our model leverages visual and motion information to produce accurate memorability scores.} Here, we compare the contributions of the frames stream (static information only) and the optical flow stream (motion information only) to separate the contributions of visual features and temporal dynamics.
    \textbf{Left:} Flow decreases the frames prediction. The frames stream detects memorable features like a human face or saturated colors, but the flow stream detects a static video and predicts lower memorability.
    \textbf{Right:} Flow increases the frames prediction. The flow stream picks up on dynamic patterns like fast bikers or a baby falling and increases the memorability prediction.
    }
    \label{fig:td_enhances_va}
\end{figure}

\noindent \textbf{3D Architectures: Video and Optical Flow.} 
Training on the RGB videos allows the network to access information on both motion and visual features, while training on optical flow lets us isolate the effects of motion. 
We train I3D architectures \cite{carreira2017} on raw video and optical flow (computed using OpenCV's TV-L1 implementation). 
Our models  were pretrained on the ImageNet and Kinetics datasets. %
We test the different visual feature architectures on both Memento10k and VideoMem videos. 
Our results are in the top section of Table \ref{tab:ablation_study}. 
Out of the three input representations (frames, flow, and video), optical flow achieves the poorest performance, probably because of the lack of access to explicit visual features like color and shape. 
Static frames perform remarkably well on their own, outperforming a 3D-video representation on VideoMem (as VideoMem is a fairly static dataset, this result is reasonable). 
Even with the relatively high level of motion in the Memento10k dataset (see Fig. \ref{fig:dyn_fig}), the video and frames streams perform comparably. 
For both datasets, combining the three streams maximizes performance, which is consistent with previous work \cite{carreira2017} and reinforces that both visual appearance and motion are relevant to predicting video memorability. 
In fact, our three-stream approach leverages motion information from the optical flow stream to refine its predictions, as shown in Fig. \ref{fig:td_enhances_va}.

\subsection{Modeling Semantic Features} \label{section:modeling-semantic-features}

It is well-known that semantics are an important contributor to memorability \cite{Cohendet2019ICCV,cohendet2018annotating,Konkle2010Concepts,shekhar2017}.
To increase our model's ability to extract semantic information, we jointly train it on memorability prediction and a captioning task, which ensures that the underlying representation learned for both problems contains relevant event semantics.
To test this approach, we enhance the video stream from the previous section with an additional module that aims to solve one of two tasks: generating captions or learning a joint text-video embedding. 
We augment the video stream as opposed to the flow or frames streams because it contains both visual and motion information required to reconstruct a video caption. 

\noindent \textbf{Caption Generation.} 
Our first approach is to predict captions directly. This has the benefit of forcing the model to encode a rich semantic representation of the image, taking into account multiple objects, actions, and colors.
However, it also involves learning an English language model, which is tangential to the task of predicting memorability. 
We feed the output features of the video I3D base into an LSTM that learns to predict the ground-truth captions.
For Memento10k, we tokenize our 5 ground-truth captions and create a vocabulary of 3,870 words that each appear at least 5 times in our training set. 
We use pre-trained FastText word embeddings \cite{bojanowski2016enriching} to map our tokens to 300-dimensional feature vectors that we feed to the recurrent module.
The VideoMem dataset provides a single brief description with each video, which we process the same way.
We train with teacher forcing and at test-time feed the output of the LSTM back into itself.

\noindent \textbf{Mapping Videos into a Semantic Embedding.} Our second approach is to learn to map videos into a sentence-level semantic embedding space using a triplet loss. 
We pre-compute sentence embeddings for the captions using the popular transformer-based network BERT \cite{bert}.\footnote{Computed using \cite{bert-as-a-service}}
At training time, we stack a fully-connected layer on top of the visual encoder's output features %
and use a triplet loss with squared distance to ensure that the embedded representation is closer to the matching caption than a randomly selected one from our dataset. 
This approach has the benefit that our network does not have to learn a language model, but it may not pick up on fine-grained semantic actors in the video. 

\noindent \textbf{Captioning Results.} 
The results of our captioning experiments are in the second section of Table \ref{tab:ablation_study}. 
Caption generation outperforms the semantic embedding approach on Memento10k. Learning to generate captions provides a boost over only the video stream for both datasets.
Fig. \ref{fig:predictions} contains examples of captions generated by our model. 

\subsection{Modeling Memorability Decay}

Up until this point, we have evaluated our memorability predictions by converting them to rankings and comparing them to the ground truth.
However, the Memento10k data and our parameterization of the memorability decay curve unlocks a richer representation of memorability, where $m_t^{(v)}$ is the true probability that an arbitrary person remembers video $v$ at lag $t$. Thus, we also investigate techniques for predicting the ground-truth values of the memorability decay curve.
Again, we consider two alternative architectures. 

\noindent \textbf{Mem-$\boldsymbol{\alpha}$ Model.} ``Mem-$\alpha$'' models produce two outputs by regressing to a video's memory score and decay coefficient. 
To train these models, we define a loss that consists of uniformly sampling 100 values along the true and predicted memorability curves and calculating the Mean Absolute Error on the resulting pairs. Equation \ref{eq1} can then be used to predict the raw hit rate at a different lag.

\noindent \textbf{Recurrent Decay Model.} 
This model directly outputs multiple probability values corresponding to different points on the decay curve. 
It works by injecting the feature vector produced by the video encoder into the hidden state of an 8-cell LSTM, where the cells represent evenly spaced lags from $t=40$ through $180$. 
At each time step, the LSTM modifies the encoded video representation, which is then fed into a multi-layer perceptron to generate the hit rate at that lag. The ground truth values used during training are calculated from $\alpha^{(v)}$ and $m^{(v)}_{80}$ using Equation \ref{eq1}.

\begin{SCtable}
    \centering
    \begin{tabular}{|c|c|ccc|}
         \hline
         \multirow{2}{*}{\textbf{Approach}} & \multirow{2}{*}{\textbf{RC}} & \multicolumn{3}{c|}{$\boldsymbol{R^2}$} \\
         & & \textbf{t=40} & \textbf{t=80} & \textbf{t=160} \\  
         \hline
         Mem-$\alpha$ & \textbf{0.604} & 0.146 & 0.227 & 0.121 \\
         Recur. head & 0.599 & \textbf{0.298} & \textbf{0.364} & \textbf{0.219} \\
         \hline
    \end{tabular}
    \caption{\textbf{Multi-lag memorability prediction}: Rank correlation (RC) and raw predictions ($R^2$) at 3 different lags ($t$, representing the number of intervening videos).}
    \label{tab:multi-lag-mem}
\end{SCtable}

\noindent \textbf{Decay Results.} 
We evaluate the models in two ways. 
First, we calculate rank correlation with ground truth, based on the memorability scores (defined as $m_{80}^{(v)}$).
We also compare their raw predictions for different values of $t$, for which we report $R^2$.  
The results are in Table \ref{tab:multi-lag-mem}.

We find that the Mem-$\alpha$ has better performance than the recurrent decay model in terms of rank correlation.
However, the recurrent decay model outperforms the Mem-$\alpha$ model at predicting the raw memorability values, and exhibits good results for low and high lags as well.
It makes sense that the performance of the Mem-$\alpha$ model falls off at lags further away from $T=80$, since any error in the prediction of alpha (the slope of the decay curve) is amplified as we extrapolate away from the reference lag. 
These two models present a trade-off between simplicity and ranking accuracy (mem-$\alpha$) and numerical accuracy along the entire decay curve (recurrent decay). Because of its relative simplicity and strong RC score, we use an Mem-$\alpha$ architecture for our final predictions.

\section{Model results}

\begin{SCfigure}
\centering
  \includegraphics[width=0.60\columnwidth]{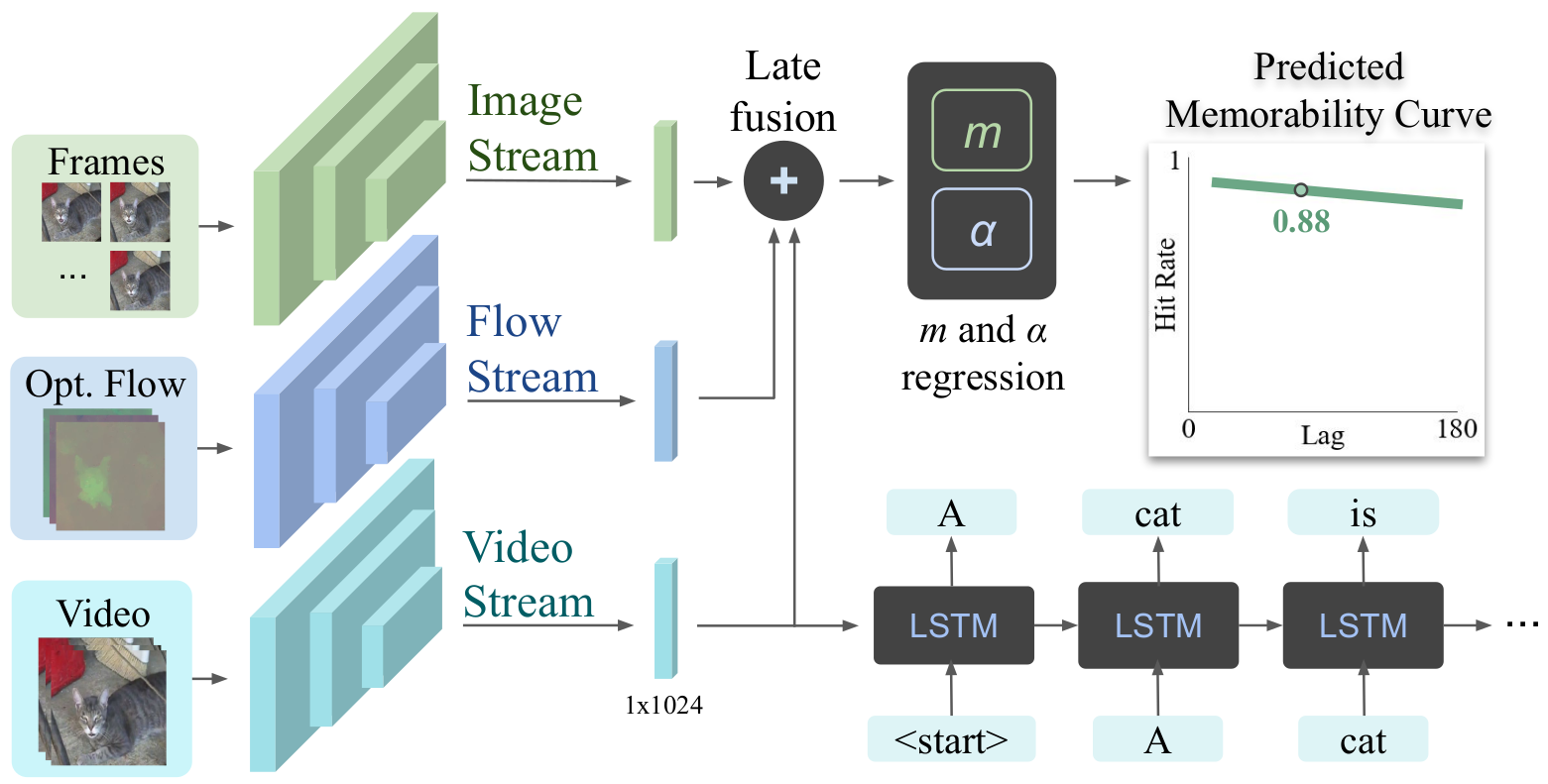}

  \caption{\textbf{The architecture of SemanticMemNet}. An I3D is jointly trained to predict memorability and semantic captions for an input video. Its memorability predictions are combined with a frames-based and optical flow stream to produce $m_{80}$ and $\alpha$, the parameters of the memorability decay curve.} ~\label{fig:semanticmemnet}
\end{SCfigure}

SemanticMemNet (Fig. \ref{fig:semanticmemnet}) combines our findings from the three previous sections. 
We use a three-stream encoder that operates on three different input representations: 1) the raw frames, 2) the entire video as a 3D unit, and 3) the 3D optical flow. 
We jointly train the video stream to output memorability scores and captions for the video. 
Each of our streams predicts both the memorability and the decay rate, which allows us to predict the probability that an observer will recall the video at an arbitrary lag within the range we studied.

\begin{figure}[h]
\centering
  \includegraphics[width = 1\columnwidth]{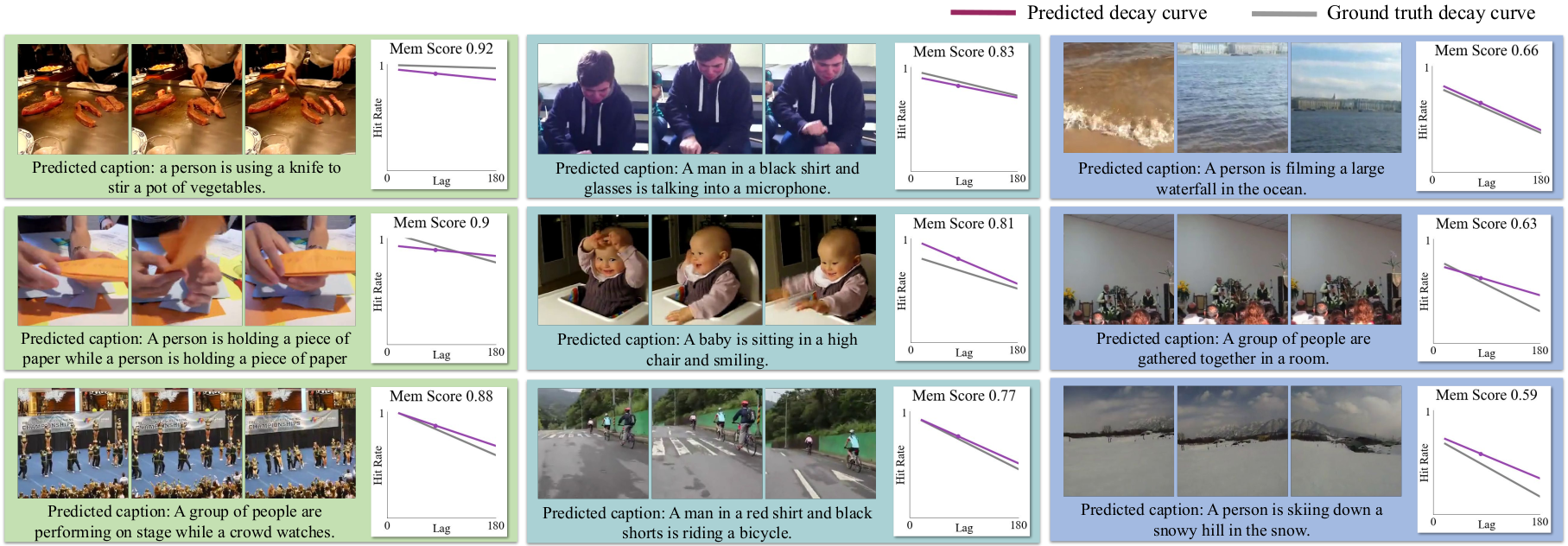}
  \caption{
  \textbf{Memorability and captions predictions from SemanticMemNet.} 
  For each example, we plot the predicted memorability decay curve based on SemanticMemNet's values 
  in purple, as well as the ground truth in gray. 
  } 
  ~\label{fig:predictions}
\end{figure}

To evaluate the effectiveness of our model, we compare against prior work in memorability prediction. MemNet \cite{khosla2015} is a strong image memorability baseline; we apply it to our videos by averaging its predictions over 7 frames uniformly sampled from the video. ``Feature extraction and regression'' is based on the approach from Shekhar et al. \cite{shekhar2017}, where semantic, spatio-temporal, saliency, and color features are first extracted from the video and then a regression is learned to predict the memorability score. The final two baselines are the best-performing models from Cohendet et al. \cite{Cohendet2019ICCV}. 
(Further details about baseline implementation can be found in the supplement.)
The results of our evaluations are in Table \ref{tab:comparison-other-work}.
Our model achieves state-of-the-art performance on both Memento10k and VideoMem. 
Example predictions generated by our model are in Fig. \ref{fig:predictions}.

\def\arraystretch{1.1} %
\begin{table}[h]
    \centering
    \begin{tabular}{|c|c|c|c|}
        \hline
         \textbf{Approach} 
         & \parbox[c]{3.0cm}{\parboxvpadding \centering  \textbf{RC - Memento10k (test set)} \parboxvpadding} 
         & \parbox[c]{2.8cm}{\parboxvpadding \centering  \textbf{RC - VideoMem (validation set\tablefootnote{We use the VideoMem validation set as the test set has not been made public.})} \parboxvpadding } \\
         \hline
         Human consistency & 0.730 & 0.616 \\ 
         \hline
         MemNet Baseline \cite{khosla2015} & \edit{0.485} & 0.425 \\ 
         Feature extraction + regression (as in \cite{shekhar2017})* & \edit{0.615} & 0.427 \\
         Cohendet et al. (ResNet3D) \cite{Cohendet2019ICCV} & \edit{0.574} & 0.508 \\
         Cohendet et al. (Semantic)\cite{Cohendet2019ICCV} & \edit{0.552} & 0.503 \\ 
         \hline
         \textbf{SemanticMemNet} & \textbf{0.663} & \textbf{0.556}  \\ 
         \hline
    \end{tabular}
    \caption{
    \textbf{Comparison to state-of-the-art} on Memento10k and VideoMem. Our approach, SemanticMemNet, approaches human consistency and outperforms previous approaches.
    {\small *Uses ground-truth captions at test-time.}
    }
    \label{tab:comparison-other-work}
\end{table}

\section{Conclusion}

\begin{figure}[t!]
\centering
  \includegraphics[width=1\columnwidth]{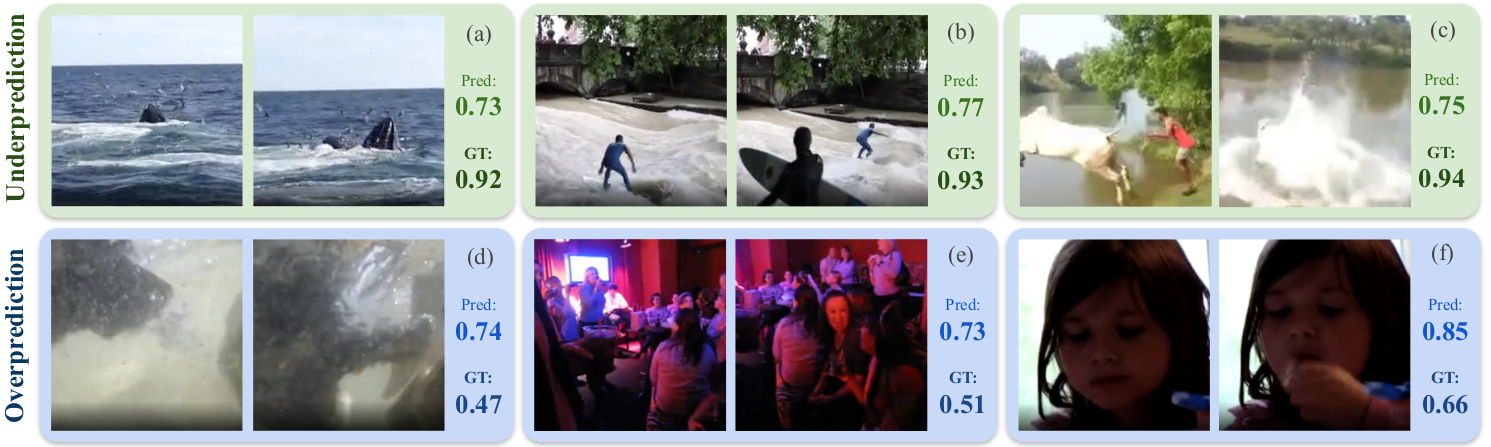}
  \caption{
    \textbf{Under and overpredictions of SemanticMemNet.} 
    Our network underestimates the memorability of visually bland scenes with a single distinctive element, 
    like a whale sighting \textbf{(a)}.
    It can fail on out-of-context scenes, like someone surfing on a flooded concrete river \textbf{(b)}, or surpising events, like a man getting dragged into a lake by a cow \textbf{(c)}.
    By contrast, it overestimates the memorability of choppy, dynamic scenes without clear semantic content \textbf{(d)} and of scenes that contain memorable elements, such as humans and faces, but that are overly cluttered \textbf{(e)}, dark \textbf{(f)}, or shaky.
  } 
  \label{fig:failure_cases}
\end{figure}

\textbf{Our contributions.} We introduce a novel task (memorability decay estimation) and a new dynamic dataset with memorability scores at different delays. 
We propose a mathematical formulation for memorability decay and a model that takes advantage of it to estimate both memorability and decay. 

\noindent \textbf{Limitations and Future Work.} 
Memorability is not a solved problem. 
Fig. \ref{fig:failure_cases} analyzes instances where our model fails because of competing visual attributes or complex semantics.
Furthermore, there is still room for improvement in modeling memorability decay (Table \ref{tab:multi-lag-mem}) and extending our understanding of memorability to longer sequences.
Our approach makes progress towards continuous memorability prediction for long videos (i.e. first-person live streams, YouTube videos) where memorability models should handle past events \textit{and} their decay rates, to assess memorability of events a    t different points in the past. 
\edit{To encourage exploration in this direction, we have released a live demo\footnote{http://demo.memento.csail.mit.edu/} of SemanticMemNet that extracts memorable segments from longer video clips.}

\noindent \textbf{The utility of memorability}. Video memorability models open the door to many exciting applications in computer vision.
They can be used to provide guidance to designers, educators, and models to generate clips that will be durable in memory. 
They can improve summarization by selecting segments likely to be retained. 
They can act as a measure of the utility of different segments in space-constrained systems; for instance, a camera in a self-driving car or a pair of virtual assistant glasses could discard data once it has fallen below a certain memorability threshold. 
Predicting visual memory will lead to systems that make intelligent decisions about what information to delete, enhance, and preserve.

\noindent \textbf{Acknowledgments.} We thank Zoya Bylinskii and Phillip Isola for their useful discussions and Alex Lascelles and Mathew Monfort for helping with the dataset.

\par\vfill\par

\clearpage
{\small
\bibliographystyle{splncs04}
\bibliography{egbib}
}

\end{document}